\documentclass{article}

\usepackage[final]{workshop}

\usepackage[utf8]{inputenc} 
\usepackage[T1]{fontenc}    
\usepackage{hyperref}       
\usepackage{url}            
\usepackage{booktabs}       
\usepackage{amsfonts}       
\usepackage{nicefrac}       
\usepackage{microtype}      
\usepackage{xcolor}         

\usepackage{makecell, multirow, tabularx}

\usepackage{amsmath,amsfonts,bm}

\def\eqref#1{equation~\ref{#1}}

\def\1{\bm{1}}

\DeclareMathAlphabet{\mathsfit}{\encodingdefault}{\sfdefault}{m}{sl}
\SetMathAlphabet{\mathsfit}{bold}{\encodingdefault}{\sfdefault}{bx}{n}

\usepackage{amsfonts} 
\usepackage{amssymb} 
\usepackage{bm} 
\usepackage{dsfont} 

\newcommand{\R}{\mathbb{R}}
\newcommand{\E}{\mathbb{E}}

\usepackage{mathtools} 
\DeclarePairedDelimiterX{\infdivx}[2]{(}{)}{%
  #1\;\delimsize\|\;#2%
}

\newcommand{\norm}[1]{\left\lVert#1\right\rVert}

\newcommand{\mbf}[1]{\mathbf{#1}}

\newcommand{\sss}{\mkern 1mu}

\DeclareMathOperator*{\argmax}{\arg\!\max}

\usepackage{transparent}

\usepackage{tikz}
\newcommand{\circo}{~\raisebox{1pt}{\tikz \draw[line width=0.4pt] circle(1.1pt);}~}
\newcommand{\ca}{A_{\sss\text{con}}}

\usepackage[figuresright]{rotating}

\title{Measuring Pre-training Data Quality without Labels for Time Series Foundation Models}

\author{%
  Songkang Wen\textsuperscript{$1$} \quad Vasilii Feofanov\thanks{Correspondence to: \href{mailto:vasilii.feofanov@huawei.com}{vasilii.feofanov@huawei.com}.}\textsuperscript{$*2$} \quad Jianfeng Zhang\textsuperscript{$1$} \\
  \textsuperscript{$1$}Huawei Noah’s Ark Lab, Shenzhen, China \quad \textsuperscript{$2$}Huawei Noah’s Ark Lab, Paris, France
}

\begin{document}

\maketitle

\begin{abstract}
Recently, there has been a growing interest in time series foundation models that generalize across different downstream tasks. A key to strong foundation models is a diverse pre-training dataset, which is particularly challenging to collect for time series classification.
In this work, we explore the performance of a contrastive-learning-based foundation model as a function of the data used for pre-training. 
We introduce contrastive accuracy, a new measure to evaluate the quality of the representation space learned by the foundation model. Our experiments reveal the positive correlation between the proposed measure and the accuracy of the model on a collection of downstream tasks. This suggests that the contrastive accuracy can serve as a criterion to search for time series datasets that can enhance the pre-training and improve thereby the foundation model's generalization.
\end{abstract}

\section{Introduction}

Nowadays, the success of large foundation models  \citep{bommasani2021opportunities} such as GPT-4~\citep{achiam2023gpt} or Llama~\citep{touvron2023llama} cause dramatic changes in the research and applications. Instead of training a tailored model for a particular task, the foundation model is pre-trained on a collection of datasets with the goal to generalize simultaneously across various downstream tasks either by fine-tuning the model or directly using its output. This workflow effectively simplifies the choice of the model architecture while reducing the requirement for amount of labeled data. 
The wave of foundation models has now reached the time series domain, including forecasting \citep{rasul2023lagllama,woo2024moirai} and classification models \citep{lin2024nutime}.

When it comes to real deployment of a time series foundation model (TSFM), a very important question is whether the pre-training dataset is sufficiently diverse, so the model generalizes well to new downstream tasks. Usually, this is verified by directly evaluating the performance on several supervised downstream tasks thereby requiring availability of annotated data samples and introducing a high cost for assesing the quality of different pre-trained datasets. Therefore, in this paper, we ask the following research question:
\begin{center}
    \textit{Can we evaluate the quality of pre-training data in an unsupervised manner?}
\end{center}

By focusing on time series classification and contrastive pre-training, we show that it is possible to leverage important information from the representation space learned by the TSFM. More precisely, if new data points are not similar to pre-training data, their embeddings will tend to not satisfy the uniformity property of contrastive learning \citep{wang2020understanding}, directly impacting the foundation model's performance. Based on this observation, we introduce a new metric called contrastive accuracy that evaluates how well spread the data points are in the embedding space. We empirically show that the proposed metric positively correlates with the model's accuracy on the tasks unseen during pre-training, allowing us to use it for measuring the quality of pre-training examples without the need to regularly test the TSFM on supervised downstream tasks.

\section{Related Work}

Time series representation learning has got a high attention in recent years, resulting in a wealth of excellent works. Many of them use the contrastive learning scheme for pre-training, including TS2Vec~\citep{yue2022ts2vec} and TS-TCC~\citep{ijcai2021-324} with a CNN backbone and TF-C~\citep{NEURIPS2022_194b8dac} with a ResNet backbone, demonstrating good performance on classification datasets, e.g, the UCR collection~\citep{dau2019ucr}. 

With the success of large language models (LLMs), popularity of the transformer architecture for time series analysis increases~\citep{Yuqietal-2023-PatchTST,ilbert2024samformer}, while the development of foundation models is becoming a prevalent task. Numerous TSFMs have been proposed recently, including Lag-Llama~\citep{rasul2023lag}, 
Time-LLM~\citep{jin2023time}, One Fits All~\citep{zhou2023one}, Brant~\citep{zhang2023brant},  
MOIRAI~\citep{woo2024moirai}.
While some of them try to adapt LLM for time series data, others train foundation  models from scratch on a large volume of time series data. In our work, we are based on the latter approach, studying how to evaluate the quality of pre-training data, which, to our knowledge, has not been addressed before.

Finally, we would like to notice that our framework reminds unlabeled performance estimation \citep{donmez2010unsupervised} where the goal is to predict model's performance on unlabeled test data. However, most of these approaches focus on single-task classification for computer vision problems \citep{hendrycks2016baseline,yu2022predicting,xie2024mano}, implying a fundamentally different methodology. To the best of our knowledge, time series foundation models have never been considered in this domain before.

\section{Proposed Method}
\label{sec:method}
First, in Section \ref{sec:problem-setup}, we introduce the setup we are working in. Then, we present briefly the architecture of our foundation model and how it was pre-trained in Section \ref{sec:architecture}, respectively. Finally, in Section \ref{sec:contrastive-acc}, we propose a new evaluation metric called the contrastive accuracy.

\subsection{Problem Setup}
\label{sec:problem-setup}
We consider the task of unsupervised pre-training where an \textit{unlabeled} pre-training set $\mathrm{X}_{\text{0}}$ is given with $m$ pre-training sequences of length $t$. The goal is to design a foundation model $F: \R^t \to \R^{d_{\text{hid}}}$ that projects any time series $\mbf{x}\in\R^t$ to a discriminative hidden space $\R^{d_{\text{hid}}}$. In this work, the quality of the foundation model is evaluated on a collection of downstream tasks $\mathcal{D}=\{D_i\}_{i=1}^{p}$, where $D_i$ consists of observations $\mathrm{X}_i$ and labels $\mathrm{Y}_i$. For each downstream task, we perform a train-test split, append $F$ with a linear head, fine-tune it on the training set and compute the accuracy score on the test set. The performance of $F$ with a pre-training dataset $\mathrm{X}_{\text{0}}$ that is averaged over all downstream tasks is denoted by $\mathcal{P}_{\text{test}}(\mathrm{X}_0)$. Similarly, $\mathcal{P}_{\text{train}}(\mathrm{X}_0)$ denotes the average performance when it is evaluated on the training set.

\subsection{Architecture and Pre-training}
\label{sec:pre-training}

Similarly to \citet{Yuqietal-2023-PatchTST} and \citet{lin2024nutime}, we use the ViT ~\citep{dosovitskiy2021vit} as the backbone, but our implementation slightly differs from theirs. Instead of dividing a sequence into disjoint patches, we employ a single CNN layer allowing them to be overlapped. Similarly to other TSFMs, we reshape any time series to a fixed sequence length equal to 512. More details on the architecture are given in Appendix \ref{sec:architecture}.

For pre-training of the foundation model, we use the contrastive learning that aims to train a such encoder that outputs similar representations for two random augmentations of the same sample (positive pair) and dissimilar representations for augmentations of two different samples (negative pair).
More formally, let $\mathcal{T}$ be a considered space of transformations (augmentations) such that $\forall\phi\in\mathcal{T},\mbf{x}\in\mathcal{X}$ we have $\phi(\mbf{x})\in\mathcal{X}$. In our experiments, we have considered the \texttt{RandomCropResize} with a crop length varying from $70\%$ to $80\%$. To measure similarity of two embeddings, we first project the output of the foundation model $F(\mbf{x})$ to a lower dimension using a MLP projector $g: \R^{d_{\text{hid}}} \to \R^{d'_{\text{hid}}}$ and then compute the cosine similarity between the two vectors defined as follows:
\begin{align*}
    s_{\cos}(\mathbf{q}, \mathbf{k}) := \frac{\mbf{q}^\top\mbf{k}}{\norm{\mbf{q}}\cdot\norm{\mbf{k}}},\qquad \forall(\mbf{q}, \mbf{k})\in\R^{2d'_{\text{hid}}}.
\end{align*}
Given a batch $B=\{\mbf{x}_i\}_{i=1}^b$, for each example $\mbf{x}_i$, we sample two augmentation functions $\phi$ and $\psi$ uniformly from $\mathcal{T}$, i.e., $\phi,\psi\sim\mathcal{U}(\mathcal{T})$, compute the pairwise similarities between all the examples in the following way:
\begin{align*}
    \mbf{s}_i(\phi, \psi) = \left[s_{\cos}\left(g\circo F\circo\phi(\mbf{x}_i),  \, g\circo F\circo\psi(\mbf{x}_j)\right)\right]_{j=1}^b \in \R^b.
\end{align*}

Following \citet{oord2018representation} as well as \citet{he2020momentum} and denoting the cross-entropy error function by $l_{\text{ce}}: \R^b\times\{1,\dots,b\}\to\R$, we update the weights of $F$ and $g$ by minimizing the contrastive loss 
defined by $\sum_{i=1}^b l_{\text{ce}}\left(\frac{\mbf{s}_i(\phi, \psi)}{T},\  i\right),$
where $T\in(0,+\infty)$ is a temperature.

\subsection{Contrastive Accuracy}
\label{sec:contrastive-acc}

It has been shown that a good representation learned by contrastive learning should satisfy uniformity property, i.e., to have a feature distribution that tends to be uniform on the unit hypersphere in order to preserve maximal information \citep{wang2020understanding}. Based on this observation, we introduce the \textit{contrastive accuracy} metric (CA, denoted by $\ca$), that measures how scattered the embeddings of the evaluation examples $\mathrm{X}'=\{\mbf{x}_i\}_{i=1}^n$ are in
the representation space. Specifically, we count how many examples have two embeddings (obtained from the two augmentations) to be the nearest neighbors with respect to the other examples in the dataset:
\begin{align*}
    \ca^{(\mathrm{X}')}(\mathrm{X}_0) &:= \E_{\phi\sim\,\mathcal{U}(\mathcal{T})}\E_{\psi\sim\,\mathcal{U}(\mathcal{T})}  \left[\frac{1}{n}\sum_{i=1}^{n} \,\mathbb{I}\left(\argmax_{j=\{1,\dots,n\}}[\mbf{s}_i(\phi, \psi)]_j  = i\right)\right],
\end{align*}
where $\mathrm{X}_0$ denotes the pre-training dataset on which the foundation model $F$ was trained. When the dataset size $n$ is large, we split the data into disjoint batches and evaluate $\mbf{s}_i(\phi, \psi)$ only on those examples that belong to the same batch as $i$. Further, we will experimentally show that the contrastive accuracy is able to hint what generalization performance on the downstream tasks we can expect from the foundation model pre-trained on $\mathrm{X}_0$.

\section{Experiments}
\label{sec:experiments}

In our experiments, we use the UCR~\citep{dau2019ucr} collection to demonstrate the benefit of the methodology proposed in Section \ref{sec:method}.
To minimize the effect of randomness, we run each experiment 5 times and report the averaged values. All experiments were performed on a single NVIDIA Tesla V100-32GB GPU card.

\subsection{Correlation with Performance}
\label{sec:subset-exp}

\begin{figure}[h]
  \centering
  \includegraphics[width=\linewidth]{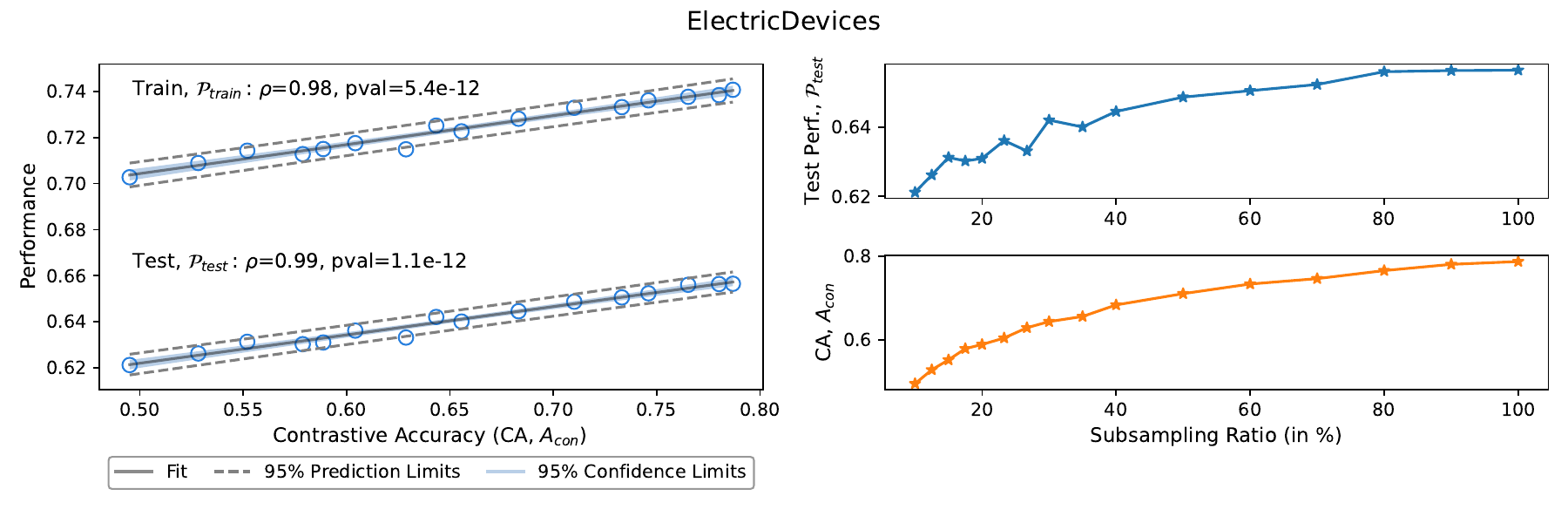}
  \caption{The correlation ($\rho$) between the contrastive accuracy and the foundation model's performance when varying the ratio of examples used for pre-training.}
  \label{eg_for_subsetSelection}
\end{figure}

In our first experiment, we show that the contrastive accuracy is able to correlate with the generalization performance of the pre-training model. Given a pre-training data $\mathrm{X}_0$, we randomly draw a subsample $\mathrm{X}^{(r\%)}_0$, which contains $r\%$ examples of $\mathrm{X}_0$, then evaluate $A_{\text{con}}^{(\mathrm{X}_0)}(\mathrm{X}^{(r\%)}_0)$ and compare it with the train and the test performance $\mathcal{P}_{\text{train}}(\mathrm{X}^{(r\%)}_0)$ and $\mathcal{P}_{\text{test}}(\mathrm{X}^{(r\%)}_0)$. As $\mathrm{X}_0$, we have picked one of the four largest datasets in the UCR (ElectricDevices, Crop, FordB, and FordA), and evaluate the performance on the rest 127 datasets in average. Figure \ref{eg_for_subsetSelection} illustrates the results for ElectricDevices when varying $r\in[10\%, 100\%]$, and the other results can be found in Appendix \ref{sec:exp-appendix}. We can observe that the contrastive accuracy correlates well with the performance, approximating its growth with increasing subsampling ratio. This allows us to perform model selection without directly testing the pre-training model on the downstream tasks. For example, this experiment may help to identify sufficient number of pre-training examples per dataset when the model is trained on a collection of different pre-training datasets.

\subsection{Improvement Prediction}
\label{dataset_selection}

In this experiment, in addition to the pre-training dataset $\mathrm{X}_0$, we consider having another one $\mathrm{X}_0^{\text{new}}$ and ask whether it is possible to predict the performance improvement from including $\mathrm{X}_0'$ to the pre-training data, i.e., $\Delta\mathcal{P}(\mathrm{X}_0, \mathrm{X}_0^{\text{new}}) :=\mathcal{P}(\mathrm{X}_0\cup \mathrm{X}_0^{\text{new}})-\mathcal{P}(\mathrm{X}_0)$. For this, we propose to consider $\Delta A_{\text{con}}(\mathrm{X}_0, \mathrm{X}_0^{\text{new}}) :=A_{\text{con}}^{(\mathrm{X}_0^{\text{new}})}(\mathrm{X}_0^{\text{new}})-A_{\text{con}}^{(\mathrm{X}_0^{\text{new}})}(\mathrm{X}_0)$ and measure its correlation with $\Delta\mathcal{P}(\mathrm{X}_0, \mathrm{X}_0^{\text{new}})$. For this experiment, we split the UCR collection into two disjoint sets denoted by $\mathcal{C}=\{\mathrm{X}_0^{(i)}\}_{i=1}^{12}$ and $\mathcal{D}=\{\mathrm{X}_i\}_{i=13}^{128}$, where the latter is used to evaluate the performance as described in Section \ref{sec:problem-setup} (more details are given in Appendix \ref{sec:exp-appendix}). 
In the first experiment, we fix the initial pre-training dataset $\mathrm{X}_0$ by taking one from $\mathcal{C}$ and vary $\mathrm{X}_0^{\text{new}}$ within $\mathcal{C}\setminus\{\mathrm{X}_0\}$. In the second experiment, we fix $\mathrm{X}_0^{\text{new}}$ and vary $\mathrm{X}_0$ in the similar way. For each pair ($\mathrm{X}_0$, $\mathrm{X}_0^{\text{new}}$, we compute  $\Delta\mathcal{P}(\mathrm{X}_0, \mathrm{X}_0^{\text{new}})$ and $\Delta A_{\text{con}}(\mathrm{X}_0, \mathrm{X}_0^{\text{new}})$ and plot the results for the two experiments in Figure \ref{eg_for_twoDatasetComp} for AllGestureWiimoteX dataset and in Appendix \ref{sec:exp-appendix} for the rest 11 datasets.
In Figure \ref{eg_for_twoDatasetComp}, we can observe a positive correlation between the difference in contrastive accuracy and the performance improvement suggesting that the proposed unsupervised criterion can help to search data that can be included to the pre-training dataset in order to improve the representaion and thereby the generalization performance of the foundation model.

\begin{figure}[h]
  \centering
  \includegraphics[width=\textwidth]{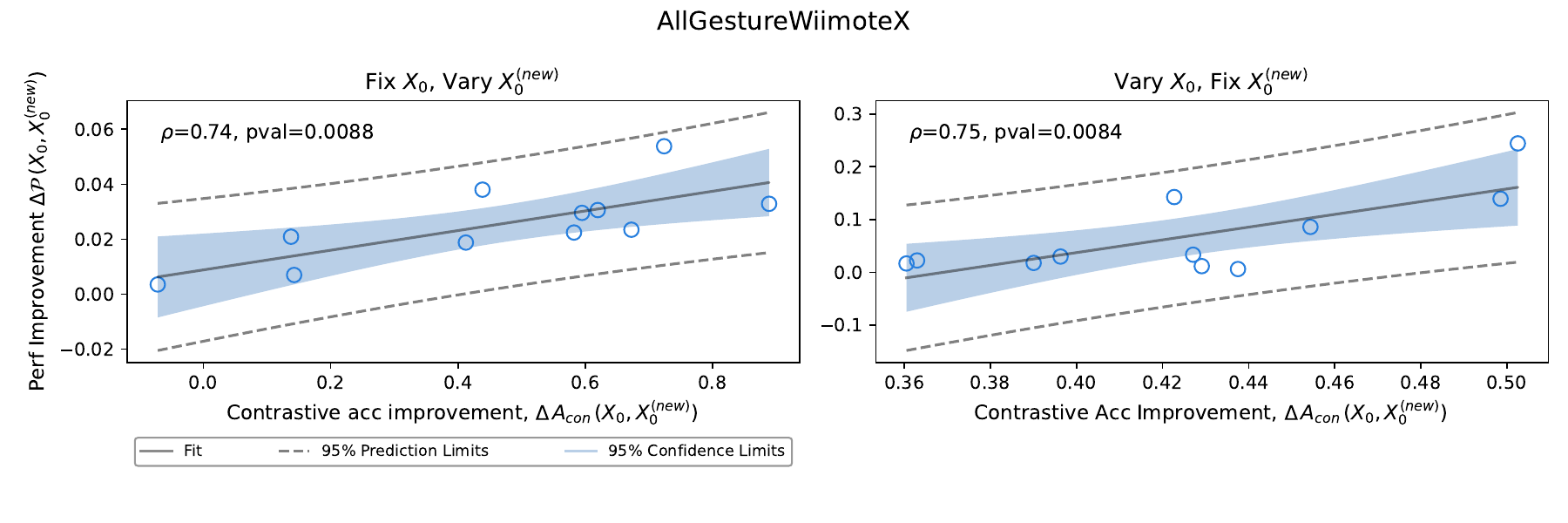}
  \caption{The correlation ($\rho$) between the improvement in contrastive accuracy and the performance improvement on 116 UCR datasets when expanding the pre-training dataset $\mathrm{X}_0$ by including $\mathrm{X}_0^{(\text{new})}$.}
    \label{eg_for_twoDatasetComp}
\end{figure}

\section{Conclusion and Future Work}

In this paper, we studied the task of evaluating the effect of pre-training data on the foundation model's performance. We proposed the contrastive accuracy and experimentally showed its promise as a criterion to select pre-training data. As a future work, we would like to test our approach with larger pre-training datasets and explore the limits of contrastive pre-training for time series data. Particularly, unlike in computer vision, there is still an open question regarding which augmentation techniques are relevant for contrastive learning in time series data and what their impact is.

\bibliography{references}
\bibliographystyle{apalike}

\appendix
\section{Architecture and Implementation Details}
\label{sec:architecture}

\begin{figure*}[h]
  \centering
  \includegraphics[width=0.8\linewidth]{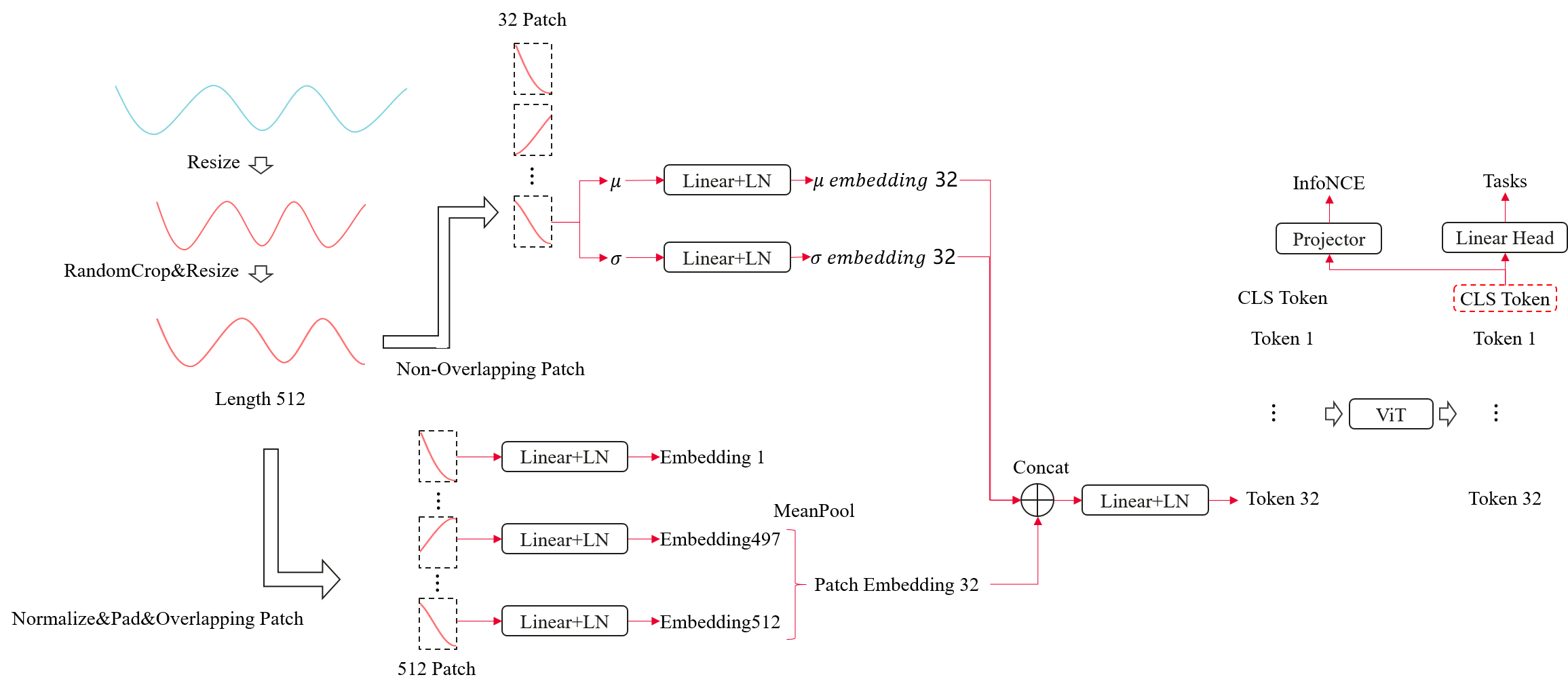}
  \caption{The framework of the model.}
    \label{model_framework}
\end{figure*}

In this paper, similarly to \citet{Yuqietal-2023-PatchTST} and \citet{lin2024nutime}, we employ the ViT (Vision Transformer) as the backbone. Our goal is to retain information effectively, so we simultaneously use both overlapping and non-overlapping patches. Here is how we achieve it: we apply a one-layer 1D-CNN to handle the overlapping patches and use mean pooling to transform their embeddings to match the number of non-overlapping patches. Next, we embed the $\mu$ and $\sigma$ values from the non-overlapping patches and concatenate the output with the overlapping patches. This result together with the class (CLS) token is then fed into the transformer. The entire network framework is depicted in Figure \ref{model_framework}.


Throughout all our experiments, we maintain the same model parameters. The chosen parameter values for different layers are outlined in Table \ref{Parameters}. To better train the model, we apply a linear learning rate warm-up in the first 10 epochs \citep{loshchilov2016sgdr} and a cosine learning rate decay in the subsequent epochs.

\begin{table}[htbp]
  \centering
  \caption{The parameters and setting for the model and experiments in this paper.}
    \begin{tabular}{cll}
    \toprule
    \multicolumn{1}{l}{Part} & Parameters & Value \\
    \midrule
    \multirow{2}[1]{*}{RandomCropResize} & Crop Scale & 0.7~0.8 \\
          & Resize Length & 512 \\
    \multicolumn{1}{l}{Non-Overlapping Patch} & Patch Length & 16 \\
    \multirow{3}[0]{*}{Overlapping Patch} & CNN Kernel Size & 17 \\
          & CNN Padding & 8 \\
          & Output Channel & 256 \\
    \multirow{7}[0]{*}{ViT} & Tokens Number & 32 \\
          & CLS Tokens & 1 \\
          & Token dimension & 256 \\
          & Layer & 6 \\
          & Number Of Head & 8 \\
          & Dimension of Head & 128 \\
          & MLP dimension & 512 \\
    \multirow{2}[0]{*}{Linear for mu} & layer & 1 \\
          & dim   & 1x32 \\
    \multirow{2}[0]{*}{Linear for std} & layer & 1 \\
          & dim   & 1x32 \\
    \multicolumn{1}{c}{InfoNCE Loss} & Temperature & 0.1 \\
    \multirow{6}[0]{*}{Non-Linear Projector} & LayerNorm &  \\
          & Layer & 2 \\
          & Input Dim & 256 \\
          & Hidden Dim & 512 \\
          & Output Dim & 256 \\
          & Activation Function & ReLU \\
    \multirow{3}[0]{*}{Classifier Linear Head} & LayerNorm &  \\
          & layer & 1 \\
          & Dim   & 256xClassNum \\
    \multirow{3}[0]{*}{Training Setting} & Learning Rate & 2e-04 \\
          & Epochs & 500 \\
          & Batch Size & 64 \\
    \multirow{3}[0]{*}{Testing Setting} & Learning Rate & 2e-04 \\
          & Epochs & 500 \\
          & Batch Size & 256 \\
    \multirow{3}[1]{*}{Optimizer} & Type  & AdamW \\
          & Betas & 0.9~0.999 \\
          & Weight Decay & 0.05 \\
    \bottomrule
    \end{tabular}%
  \label{Parameters}%
\end{table}%

\section{Experiments}
\label{sec:exp-appendix}

\begin{figure}[ht!]
    \centering
    \includegraphics[width=\textwidth]{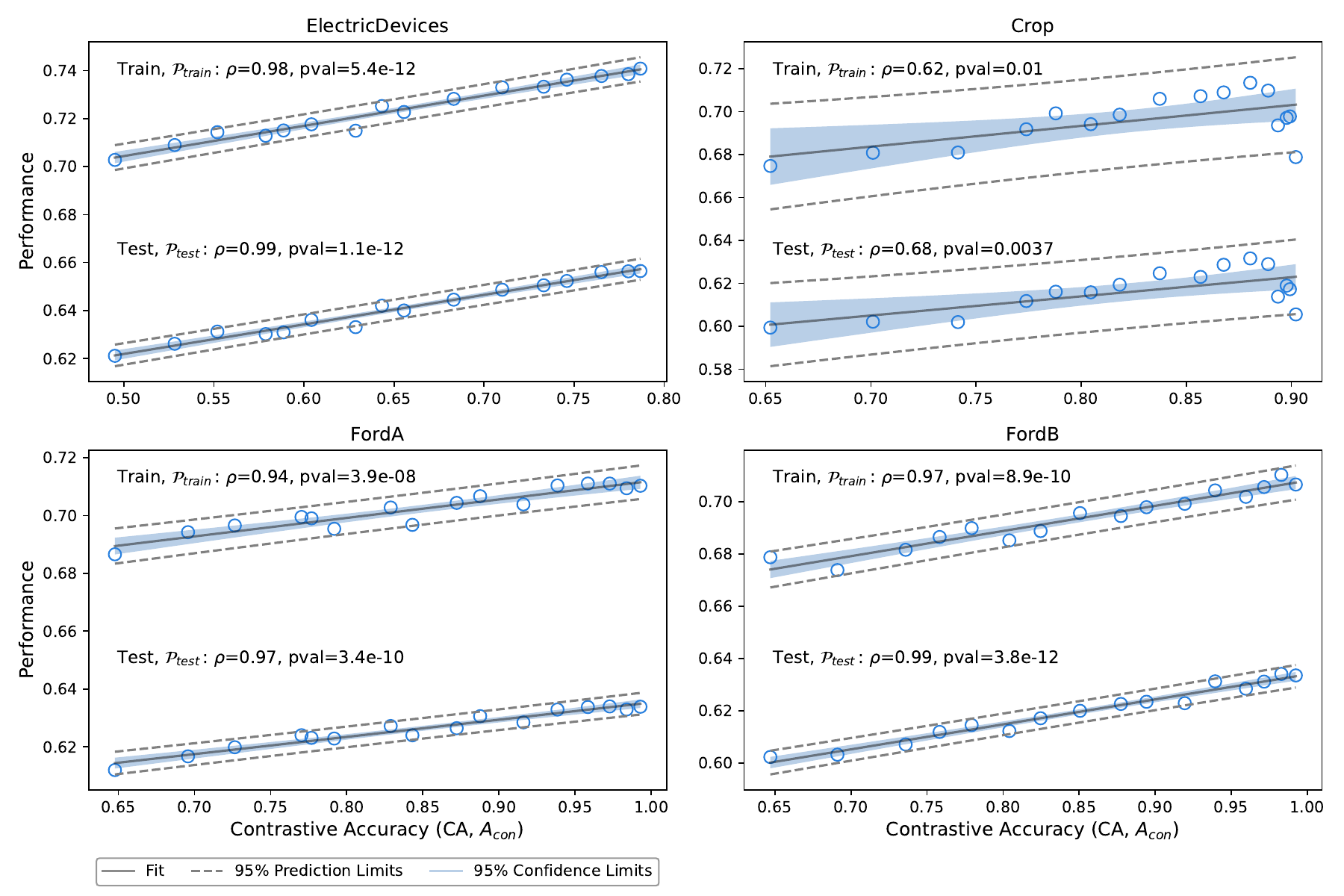}
    \caption{The correlation ($\rho$) between the contrastive accuracy and the foundation model's performance when varying the ratio of examples used for pre-training.}
    \label{fig:correlation-4datasets}
\end{figure}

In Figure \ref{fig:correlation-4datasets}, we display the complete results for the experiment presented in Section \ref{sec:subset-exp}.

\begin{figure}[ht!]
    \centering
    \includegraphics[width=\textwidth]{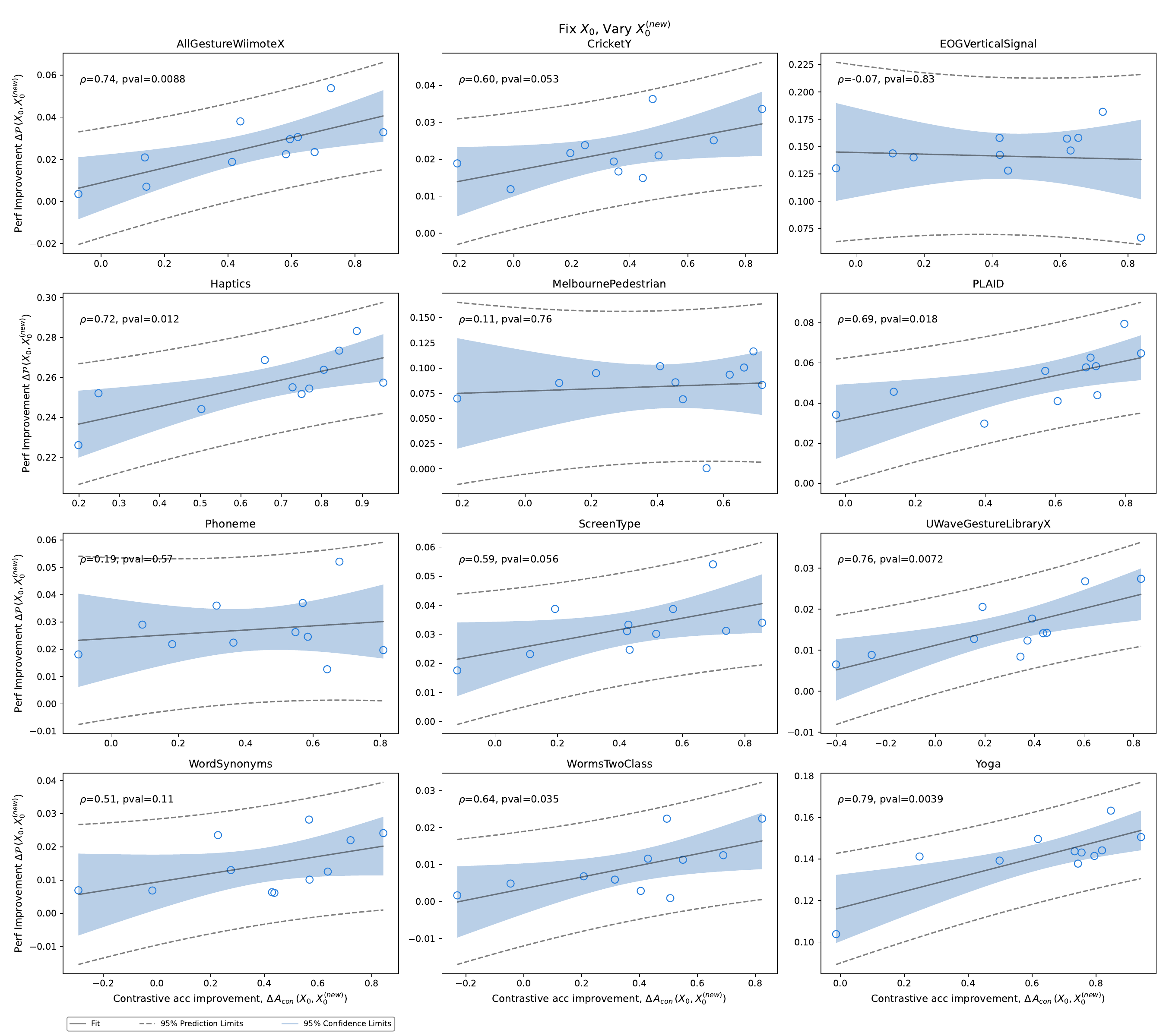}
    \caption{The correlation ($\rho$) between the improvement in contrastive accuracy and the performance improvement on 116 UCR datasets when expanding the pre-training dataset $\mathrm{X}_0$ by including $\mathrm{X}_0^{(\text{new})}$. To measure correlation, we fix $\mathrm{X}_0$ and vary $\mathrm{X}_0^{(\text{new})}$ across 11 datasets.}
    \label{fig:fix-x0-vary-x0-new}
\end{figure}

\begin{figure}[ht!]
    \centering
    \includegraphics[width=\textwidth]{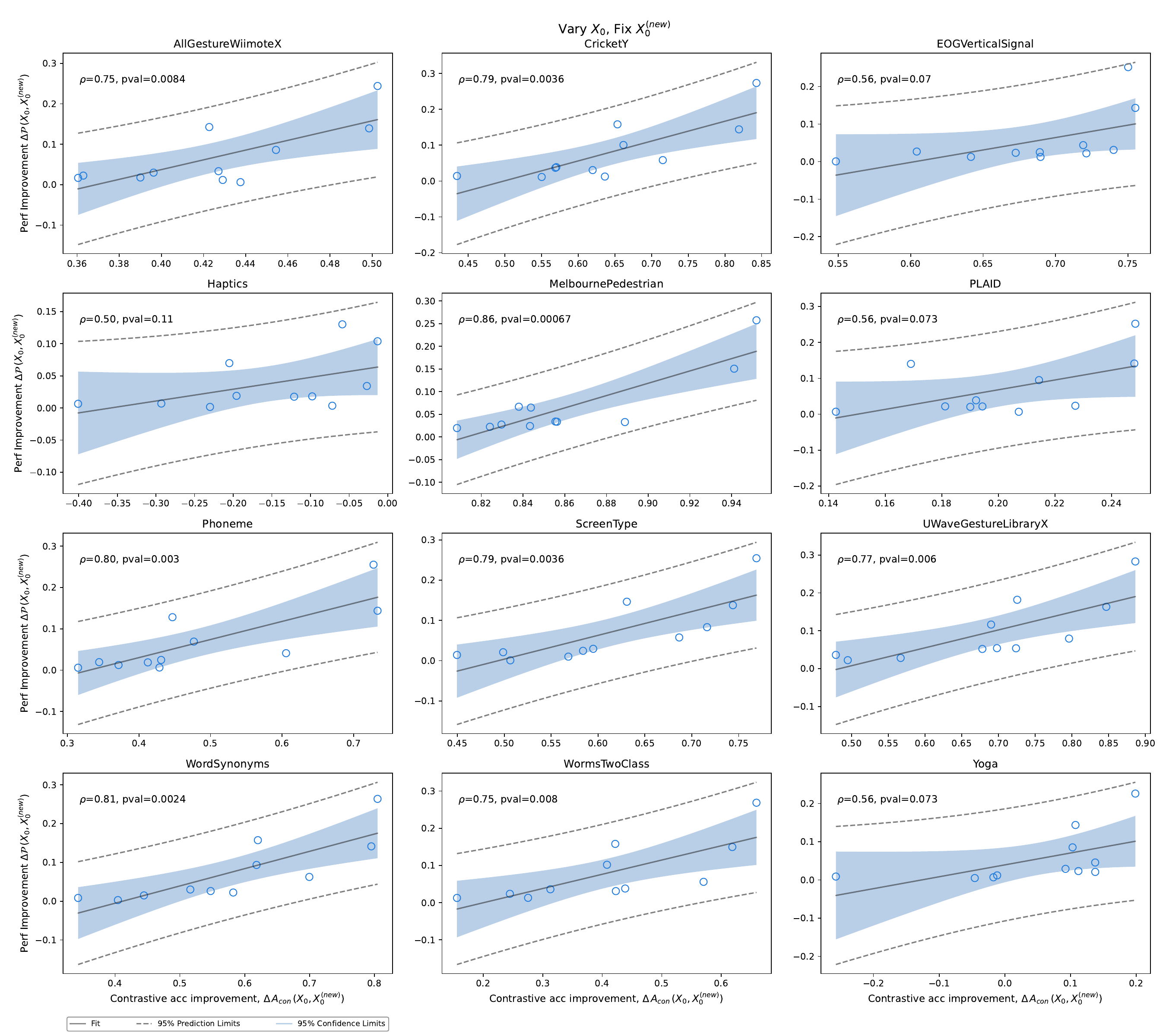}
    \caption{The correlation ($\rho$) between the improvement in contrastive accuracy and the performance improvement on 116 UCR datasets when expanding the pre-training dataset $\mathrm{X}_0$ by including $\mathrm{X}_0^{(\text{new})}$. To measure correlation, we fix $\mathrm{X}_0^{(\text{new})}$ and vary $\mathrm{X}_0$ across 11 datasets.}
    \label{fig:vary-x0-fix-x0-new}
\end{figure}

Next, in Figure \ref{fig:fix-x0-vary-x0-new} and \ref{fig:vary-x0-fix-x0-new}, we give the complete results of the experiment described in Section \ref{dataset_selection}. For the pre-training collection $\mathcal{C}$, we chose the following 12 UCR datasets: AllGestureWiimoteX, CricketY, EOGVerticalSignal, Haptics, MelbournePedestrian, PLAID, Phoneme, ScreenType, UWaveGestureLibraryX, WordSynonyms, WormsTwoClass,  Yoga.

\end{document}